\documentclass[conference]{IEEEtran}
\IEEEoverridecommandlockouts

\usepackage[T1]{fontenc}
\usepackage[utf8]{inputenc}
\usepackage{cite}
\usepackage{amsmath,amssymb,amsfonts}
\usepackage{graphicx}
\usepackage{textcomp}
\usepackage{xcolor}
\usepackage{booktabs}
\usepackage{array}
\graphicspath{{figures/}}

\def\BibTeX{{\rm B\kern-.05em{\sc i\kern-.025em b}\kern-.08em
    T\kern-.1667em\lower.7ex\hbox{E}\kern-.125emX}}

\begin{document}

\title{Multi-Sensor Fusion for UAV Classification Based on Feature Maps of Image and Radar Data}

\newcommand{\authorcell}[2]{%
\begin{minipage}[t]{0.46\textwidth}\centering
{\normalsize #1}\\[0.4ex]
\itshape\small #2
\end{minipage}}
\author{%
\begin{minipage}{\textwidth}\centering
\authorcell{Nikos Sakellariou}{Information Technologies Institute \\
Centre for Research \& Technology -- Hellas\\
Thessaloniki, Greece \\
\upshape saknikolaos@gmail.com}\hspace{2.5em}%
\authorcell{Antonios Lalas}{Information Technologies Institute \\
Centre for Research \& Technology -- Hellas\\
Thessaloniki, Greece}\\[3ex]
\authorcell{Konstantinos Votis}{Information Technologies Institute \\
Centre for Research \& Technology -- Hellas\\
Thessaloniki, Greece}\hspace{2.5em}%
\authorcell{Dimitrios Tzovaras}{Information Technologies Institute \\
Centre for Research \& Technology -- Hellas\\
Thessaloniki, Greece}
\end{minipage}
}

\maketitle
\renewcommand{\thefootnote}{}\footnotetext{\copyright~2026 IEEE. Personal use of this material is permitted. Permission from IEEE must be obtained for all other uses, in any current or future media, including reprinting/republishing this material for advertising or promotional purposes, creating new collective works, for resale or redistribution to servers or lists, or reuse of any copyrighted component of this work in other works. Published in: \emph{2026 International Symposium on Networks, Computers and Communications (ISNCC)}, Bristol, UK, 8--10 Sept.\ 2026. DOI: 10.1109/ISNCC70543.2026.11693863}\renewcommand{\thefootnote}{\arabic{footnote}}

\begin{abstract}
The cost, flexibility, and efficiency of modern UAVs make them attractive across many applications, but their proliferation has driven a rising number of malicious or accidental incidents, making UAV detection and classification mechanisms essential. Individual sensing modalities each present complementary limitations, and existing detection systems typically rely on a single sensor or fuse modalities only at the decision level, leaving the feature-level fusion of heterogeneous image and radar detectors largely unexplored. We propose a deep neural network that fuses high-level features extracted from the individual object-detection and classification models of thermal, optronic, and radar sensors. A CNN-based architecture combines the three modalities by stacking the thermal and optronic image features along the channel axis prior to fusion with the radar features. Evaluated on a real-world multi-sensor dataset, the proposed three-modality fusion model attains an F1-score of 0.95, compared to 0.93 for the dual-modality (thermal--optronic) configuration and 0.91 for the best-performing single-sensor (thermal) baseline, confirming that fusing complementary sensor features yields measurable gains in UAV classification performance.
\end{abstract}


\section{Introduction}
Unmanned Aerial Vehicles (UAVs) have permeated modern society across civil and military domains, including industrial inspection, public-space monitoring, cartography, wildlife surveying, search and rescue, and delivery of goods. Driven by regulatory progress and sustained investment, the global market value is projected to reach \$70 billion by 2029. This rapid growth, however, has been accompanied by an ever-increasing number of accidental or malicious incidents, rendering the development of UAV detection and classification mechanisms essential.

On the other hand, airborne object detection and identification is a difficult task, particularly in the case of identifying a UAV against other airborne elements such as birds, clouds and airplanes. Traditional UAV detection methods employ systems with a single sensor modality including radar, camera and RF sensors or multiple modalities of the same type such as multiple acoustic arrays. However, unimodal sensors alone may result in poor detection range and classification accuracy. For that reason, multimodal sensor fusion systems can increase performance by replenishing missing information with another sensor's perspective of the environment \cite{ref1}. To position the proposed system within the existing literature, the remainder of this section reviews prior UAV detection and classification approaches grouped by sensing strategy: single-modality systems, then multi-modality systems, before presenting our proposed solution.

\subsection{Single modality}
Single-sensor UAV detection is typically organized into detection, verification, and classification stages. For radar, Continuous-Wave and Frequency-Modulated Continuous-Wave systems are the most cost-efficient detectors \cite{ref2}, while classification relies on machine-learning methods such as SVMs \cite{ref3}, random forests \cite{ref4}, nearest-neighbour \cite{ref5}, and increasingly deep neural networks \cite{ref6,ref7,ref8}. Recent CNN-based approaches operate on X-band FMCW surveillance data \cite{ref9} or on micro-Doppler spectrograms \cite{ref10}, and LSTM models exploit micro-Doppler signatures for joint detection, classification, and localization \cite{ref11}. Radar nonetheless struggles with UAVs flying at low altitude and presenting a small radar cross-section.

Optical detectors built on CNNs distinguish drones from birds and background \cite{ref12,ref14,ref15} and have been deployed with YOLO-based pipelines on quad-rotor imagery \cite{ref16} and 4K urban-sky monitoring \cite{ref17}, although single-camera range is limited to roughly 150\,m. Thermal sensing operates in the non-visible spectrum \cite{ref18} and has been applied to night-time UAV detection \cite{ref19,ref20} and, more broadly, to thermal object recognition and tracking with CNN and nearest-neighbour classifiers \cite{ref21,ref22}. Further modalities include acoustic arrays \cite{ref23,ref24}, 3D LADAR \cite{ref25}, and WiFi fingerprinting \cite{ref26}; however, acoustic and optical sensors are typically limited to a few hundred metres, while RF- and WiFi-based methods require the pilot to be nearby.

\subsection{Multiple modalities}
Multi-sensor systems compensate for the weaknesses of one modality with the strengths of another. Wang et al. \cite{ref27} fuse visible and thermal cameras through a CNN with Cycle-GAN thermal-image augmentation, while Svanstr\"{o}m et al. \cite{ref28} combine an ADS-B receiver, visible and thermal cameras, audio, and a fish-eye director camera, and also study detection performance versus sensor-to-target distance. Other work fuses a wide-angle and a rotating camera for multi-frame drone detection \cite{ref29}, exploits multi-dimensional RF/CSI signal features \cite{ref30}, or evaluates per-sensor and decision-level fusion of visible, thermal, and acoustic detectors \cite{ref31}. These systems, however, fuse modalities predominantly at the decision level rather than at the feature level.

\subsection{Proposed Solution}
In this work, the drone detection task was divided into two separate sub-problems. First, a flying object is detected and localized; second, the object is classified into one of two classes: a UAV or a false alarm (birds, humans, other objects, noise). We focus on the second problem by proposing a deep learning multi-sensor fusion model for aerial object classification based on feature maps of the individual sensors' deep neural network (DNN) architectures, and introduce a novel method for fusing visible and thermal data by merging them into a new vector through feature stacking.

The use of feature maps as a generic representation was inspired by off-the-shelf CNN-feature transfer \cite{ref32,ref33} and by late-fusion schemes that combine high-level features from multiple deep architectures; Akilan et al. \cite{ref34,ref35}, for example, report a 2\%--8\% accuracy gain over unimodal baselines. Feature-level and late fusion of deep representations remains an active and effective strategy for multi-sensor UAV recognition: Lee et al. \cite{ref39} fuse CNN predictions from optical, radar, and audio sensors for up to a 15.6\% improvement over the best single sensor, Frid et al. \cite{ref40} fuse RF and acoustic features in a deep network, and single-sensor detectors continue to advance through richer multi-scale feature aggregation \cite{ref41}.

Unlike prior fusion approaches that combine only image modalities \cite{ref27}, operate on a different sensor set \cite{ref28}, or fuse feature maps from multiple CNNs trained on the same image data \cite{ref34}, the present work performs feature-map-level fusion across heterogeneous image (thermal and optronic) and 2D-radar detectors. Its main contributions are:
\begin{itemize}
\item[(i)] a late-fusion DNN that combines high-level feature maps from three heterogeneous sensors (thermal, optronic, and 2D radar) into a single CNN-based UAV-versus-false-alarm classifier;
\item[(ii)] a feature-stacking scheme that merges the thermal and optronic feature maps along the channel axis into a unified $(7,7,1536)$ tensor prior to fusion, rather than conventional flat-vector concatenation;
\item[(iii)] a temporal registration pipeline that aligns image-based and radar detections into matched multi-modal samples despite their differing acquisition rates;
\item[(iv)] a controlled single $\rightarrow$ two $\rightarrow$ three modality comparison on a real, field-collected dataset spanning varied operating conditions (day and night, ranges up to ${\sim}1.3$\,km), showing a monotonic F1-score improvement of $0.91 \rightarrow 0.93 \rightarrow 0.95$.
\end{itemize}

The rest of the paper is structured as follows: Section~\ref{sec:methods} considers the materials and methods. Section~\ref{sec:results} presents the experimental results and a discussion of the proposed methodology. Finally, Section~\ref{sec:conclusion} concludes the paper.

\begin{figure*}[t]
\centering
\includegraphics[width=0.92\textwidth]{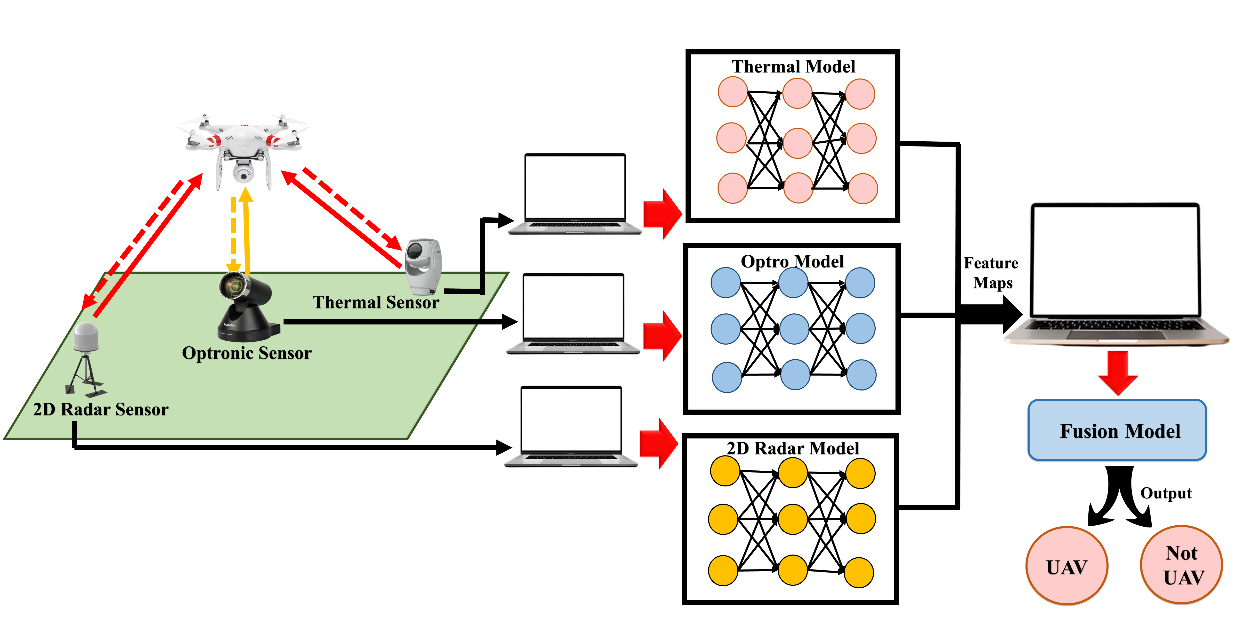}
\caption{Data flow of the complete system.}
\label{fig:dataflow}
\end{figure*}

\section{Material and Methods}\label{sec:methods}
A deep learning fusion method is introduced that combines data retrieved from three modalities (thermal, optronic and 2D radar) that incorporates different DNN architectures towards the classification of UAVs. In the terminology used throughout this paper, the thermal and optronic (electro-optical) sensors constitute the two image-based modalities, since each produces a two-dimensional pixel image, whereas the 2D radar provides a complementary, non-image modality. The data are geospatially aligned and they are fed to a separate fusion neural network that aims to binary classify captured flying objects between two classes, ``UAV'' or ``No UAV'' and hence increase the performance of the combined system against the separate models alone.

The data were collected by setting up the sensors in an open space and executing scenarios with flying UAVs. The development methodology began by extracting the available data from independent deep learning models of each sensor that were trained to localize and classify flying objects. The data are feature maps that are extracted from intermediate layers of the DNN architectures of each modality (the individual architectures are described in more detail later in this section). A CNN based architecture for the thermal features was created as a base model. Two more models were created and optimized, a fusion of thermal and optronic data and a three-modality fusion of thermal, optronic and radar data. The resulting models' performance is compared in terms of accuracy, precision, recall and F1-score to discover the optimum model for the complete system.

\begin{figure*}[t]
\centering
\includegraphics[width=0.92\textwidth]{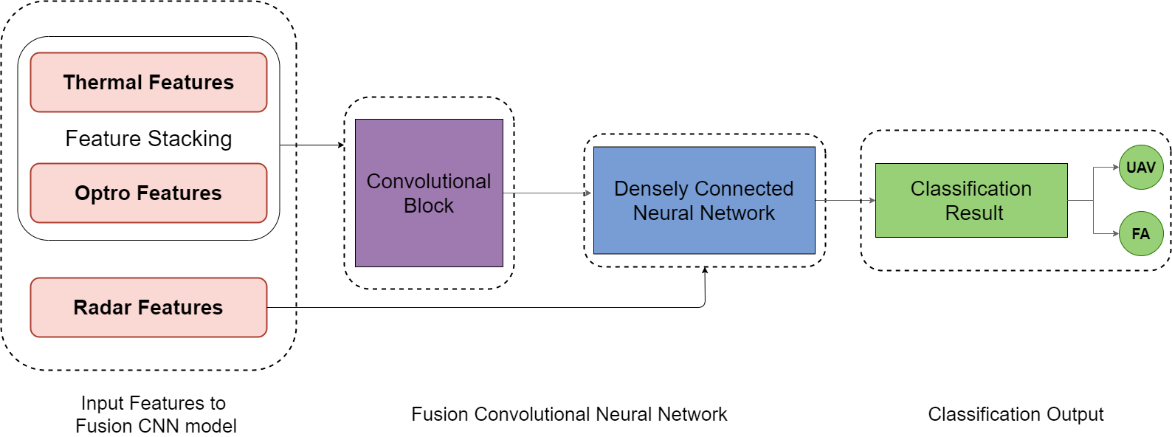}
\caption{The proposed architecture of the fusion system.}
\label{fig:fusionsystem}
\end{figure*}

Figures~\ref{fig:dataflow} and \ref{fig:fusionsystem} together serve as a flowchart of the complete pipeline and summarize how the heterogeneous sensor data are processed end to end. As shown in Fig.~\ref{fig:dataflow}, each sensor (thermal, optronic, and 2D radar) independently captures and forwards its raw data to a dedicated, pretrained deep neural network that detects and classifies candidate objects. Rather than fusing raw signals --- which differ in dimensionality, sampling rate, and physical meaning across the three modalities --- the system extracts a fixed-size high-level feature map from an intermediate layer of each sensor's network: $(7,7,1024)$ for the thermal branch, $(7,7,512)$ for the optronic branch, and a 1664-element vector for the radar branch. These feature maps are temporally registered, the thermal and optronic maps are stacked along the channel axis into a single $(7,7,1536)$ tensor, and the result is fused with the radar feature vector in the binary classification network of Fig.~\ref{fig:fusionsystem}. Operating at the feature level, rather than on raw heterogeneous signals, is what makes the three modalities compatible for joint learning.

\subsection{Data Acquisition and Processing}
The data capturing sessions took place in a town of Greece named Markopoulo in an open area. Three main scenarios were employed during data recording. A UAV coming from afar, a UAV taking off from the nearby ground and heading to a protected area and a swarm of UAVs moving towards and away from the sensor area. The capturing sessions lasted for three days, recording UAV flights during different times of day and night in clear and cloudy sky. The sensors were set at the center of the area and the operation of the drones was performed from two points: near the sensor equipment and from a location 350m away from the sensors. Multiple DJI Mavick Pro, a single DJI Matrice 210, and a single DJI Phantom 4 were used. The UAVs' maximum speed was up to 15\,m/s, their distance from the sensors reached up to 1300\,m, and their maximum altitude was 100\,m above the sensor level. Labeling was performed manually: for the optical and thermal sensors a frame per second was extracted and annotated with a bounding box, image coordinates, and a class label, while the radar data were aligned with the UAV's GPS track in the time domain to form the ground truth.

The feature maps \cite{ref42} from the independent classifiers are extracted and they are sent to the main fusion algorithm as input data. The radar module receives as input a list of detections and radar features that are stemmed from radar signal processing algorithms. The annotation includes ground truth labels for drones and false alarms. The optronic and thermal modules are trained on the annotated data and feature maps from the last layer of their networks are extracted from their respective DNN models.

\subsubsection{Radar Data}
The radar sensor is an X-band LFMCW surveillance radar (9.35\,GHz, 4\,W, PRF 3.3\,kHz) with a detection range up to 4\,km. Its raw quadrature/in-phase data are converted to Range Profile Matrices \cite{ref43} through calibration, radar-equation correction, and FFT, yielding Range-Doppler detections. From each detection, a Range-Profile signature and a set of radar features (amplitude, amplitude-to-mean ratio, and radial velocity) feed a two-stream CNN classifier whose concatenation layer produces a 1664-element feature map; the complete pipeline is detailed in \cite{ref9}. In this work the network is reused as a pretrained feature extractor: we take its 1664-element feature map as the radar input to the fusion model. The contribution of this paper therefore lies not in the individual sensor pipelines, which are adopted from prior work, but in the way their heterogeneous feature maps are temporally aligned, stacked, and fused, as detailed in the following subsections.

\subsubsection{Optronic Data}
The optronic sensor is a PTZ camera, operator-directed toward possible detections, that captures $1920 \times 1080$ video at 25\,fps. The frames feed a Faster R-CNN detector with a MobileNet \cite{ref44} backbone pre-trained on ImageNet \cite{ref45}: the base CNN produces an intermediate feature volume, a Region Proposal Network suggests 64 candidate boxes, and region cropping and pooling yield $7 \times 7 \times 512$ feature maps that are classified as UAV or background. Maps classified as UAV detections are passed to the fusion model.

\subsubsection{Thermal Data}
The thermal sensor is a Spynel camera that continuously provides a $360^{\circ}$ coverage of the area. The sensor captures video and images of size $12288 \times 640$ pixels and they are extracted at a rate of 2\,Hz to be fed to a Faster R-CNN \cite{ref46} object detection and classification model that is running on a laptop connected to the sensor. The model may detect more than one occurrence of objects in a frame and it classifies them as UAVs or background. A modified version of ResNet-50 is used as the base CNN of the Faster R-CNN architecture designed to handle large high resolution images rapidly. The feature maps that are used in the fusion model were extracted with the same method as in the case of the visible sensor and they are represented by a vector of size $7 \times 7 \times 1024$.

\subsection{Preprocessing and Data Flow}
The fusion model incorporates data received from the three modalities occupying a total of 54.8 GB of hard disk storage. The data were stored in NumPy arrays (.npz files) that contain the feature maps and their annotation information (ground truth labels, timestamps and metadata). There is a total of 29 recordings for each sensor. A python script loads and processes them one by one aiming to register the data samples of the three sensors. This process breaks down into two tasks: the feature matching between the thermal and optronic data and the registration between radar and the already matched two-modality data. The data shape of a single instance is $(7,7,512)$ for the optronic modality and $(7,7,1024)$ for the thermal modality. Each feature map corresponds to a detection from an extracted frame of the captured video. Both the optronic and infrared frames are extracted at the same time. Therefore, their feature maps were easily matched in the time domain with high alignment accuracy. The vectors from these two sensors are then merged into a new feature vector of shape $(7,7,1536)$ by stacking them across the last axis of the three-dimensional array.

The resulting samples are then matched with the corresponding radar recordings in the time domain using a one-second window ($\pm0.5$\,s around each thermal--optronic instance). This window reconciles the differing sampling rates of the radar and electro-optical sensors: it is wide enough that a radar detection is available for most thermal--optronic instances, yet narrow enough to bound misalignment --- at the observed UAV speeds (up to 14\,m/s), $\pm0.5$\,s corresponds to a displacement of at most ${\sim}7$\,m, negligible relative to the operating ranges of hundreds of metres to one kilometre. For each thermal--optronic instance, the closest radar sample with the same label is selected, forming a two-element input tuple for the fusion model: a $(7,7,1536)$ stacked image feature map and the 1664-element radar feature vector.

\begin{figure*}[t]
\centering
\includegraphics[width=0.92\textwidth]{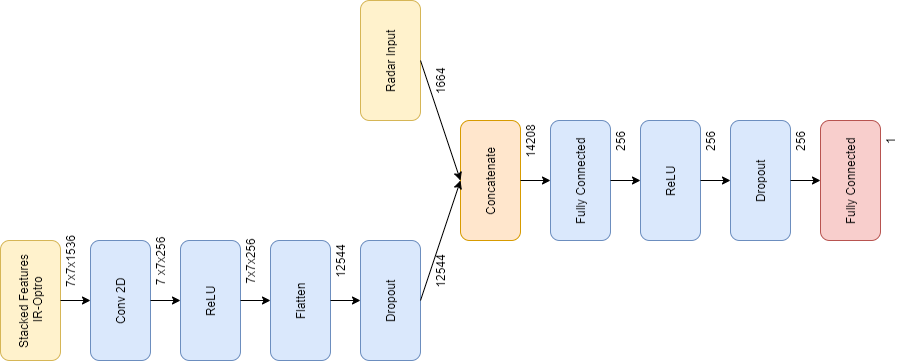}
\caption{Fusion architecture for 2D radar -- thermal -- optronic modalities.}
\label{fig:fusionarch}
\end{figure*}

\subsection{Fusion Method}
Three CNN based architectures were developed in total that utilized feature maps from the independent object detection and classification models of each sensor. An architecture relying purely on thermal data, an architecture based on fusion of thermal and optronic data and the proposed three modality architecture. The three models were compared to better measure the impact of the multimodal architecture. All the architectures were based on convolutional neural networks and were evaluated on the same testing set.

\subsubsection{Fusion Architecture}
The model was implemented in Keras/TensorFlow as a two-input binary classifier. The first input is the $(7,7,1536)$ stacked thermal--optronic feature map; the second is the 1664-element radar vector. The image input passes through a 2D convolutional layer (512 filters, $3\times3$, ReLU), whose flattened output (dropout 0.5) is concatenated with the radar input into a 14208-element vector feeding a 512-neuron fully connected layer (ReLU, dropout 0.5) and a sigmoid output, with a 0.5 threshold separating UAV from false alarm (Fig.~\ref{fig:fusionarch}). Binary cross-entropy is used as the loss, and the 7{,}189{,}506 trainable parameters are optimized with RMSprop \cite{ref47} (learning rate 0.0001, decay $10^{-7}$). This deliberately compact head --- a single convolutional layer followed by one fully connected layer with dropout --- was preferred over a deeper network because the fusion model already operates on high-level features and is trained on a limited number of synchronized samples, providing sufficient capacity to learn cross-modal correlations while limiting overfitting. In preliminary experiments, classical classifiers (SVM, random forest, XGBoost) applied to the fused features performed markedly worse, as they require the stacked maps to be vectorized, discarding the spatial and cross-modal correlations that the convolutional head exploits.

Two feature maps that share a common spatial resolution can be combined in two ways: by \emph{channel-axis stacking}, concatenating them along the depth dimension into a $(7,7,1536)$ tensor that retains the $7\times7$ spatial grid, or by \emph{flat-vector concatenation}, flattening each map and joining the two long vectors before a dense layer. We adopt the former. Because the thermal and optronic detections are temporally matched and ROI-pooled to a common $7\times7$ grid, a given spatial cell refers to approximately the same region of the detected object in both modalities. Channel stacking preserves this correspondence, so the subsequent 2D convolution applies shared $3\times3$ filters that mix thermal and optronic channels at each co-located cell and learn cross-modal spatial patterns (for example, a thermally hot region coinciding with a compact optical signature). Flat concatenation instead discards the spatial grid: a dense layer over the ${\sim}75{,}000$-element flattened vector treats every activation independently, cannot share weights across positions, and introduces far more parameters --- undesirable given the limited number of synchronized multi-sensor samples. The radar features carry no image-plane geometry and are therefore concatenated as a flat 1664-element vector \emph{after} the convolutional stage, where spatial structure is no longer required.

The two-modality model is identical but omits the radar input and concatenation layers, operating on the $(7,7,1536)$ stacked thermal--optronic input alone; the single-modality model further reduces the input to the $(7,7,1024)$ thermal tensor, with the remaining layers unchanged.

\subsubsection{Training}
Training was performed on two NVIDIA Tesla K40m GPUs (12\,GB each). Because the compact fusion head operates on pre-extracted rather than raw sensor data, training converged in about 17\,minutes---roughly 60\,s per iteration---while using 27\,GB of RAM. Per-sample inference is correspondingly lightweight, so at deployment the dominant latency is the $\pm0.5$\,s temporal-registration window rather than the network itself. The available data consist of 29 recordings: 26 were used for training and validation (split 80\%/20\%) and three were held out as an unseen test set. Crucially, the 29 recordings are data-capture sessions rather than individual examples --- each contains numerous synchronized multi-sensor detections --- so the fusion model is trained and validated on 17{,}633 individually labelled feature-map samples. Generalizability is further promoted by dropout regularization and early stopping on the validation loss, while the three held-out recordings provide an unbiased estimate of performance.

The network's hyper-parameters were tuned by fitting the network many times on different parameter combinations. The following parameters were tested with different values: learning rate, batch size, number of convolutional nodes, number of dense nodes, and activation function. The optimal hyper-parameters for the model applied on the dataset were found to be \texttt{lr=0.0001}, \texttt{batch\_size=12}, \texttt{convolutional\_nodes=512}, \texttt{dense\_nodes=512} and \texttt{activation\_function=ReLU}. The model was configured to run for 160 epochs. Early stopping was applied with the patience parameter being set to 10 epochs monitoring the validation loss. The training set was comprised of 13224 samples and the validation set of 4409.

\begin{figure}[t]
\centering
\includegraphics[width=0.49\columnwidth]{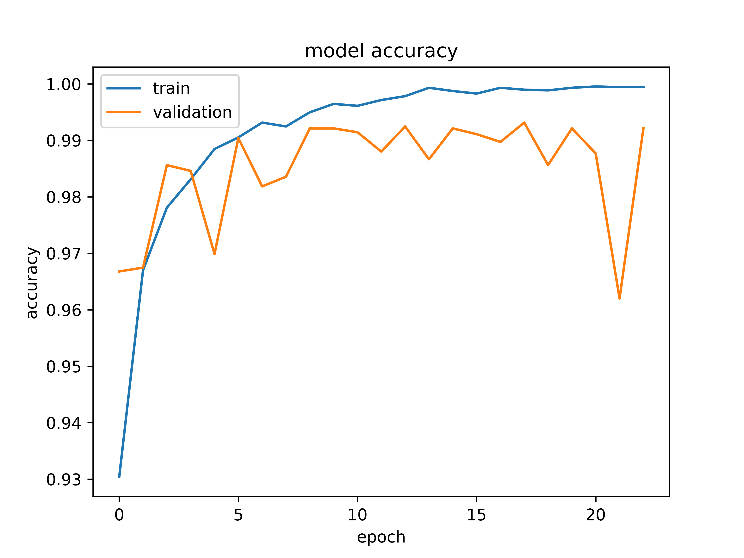}\hfill
\includegraphics[width=0.49\columnwidth]{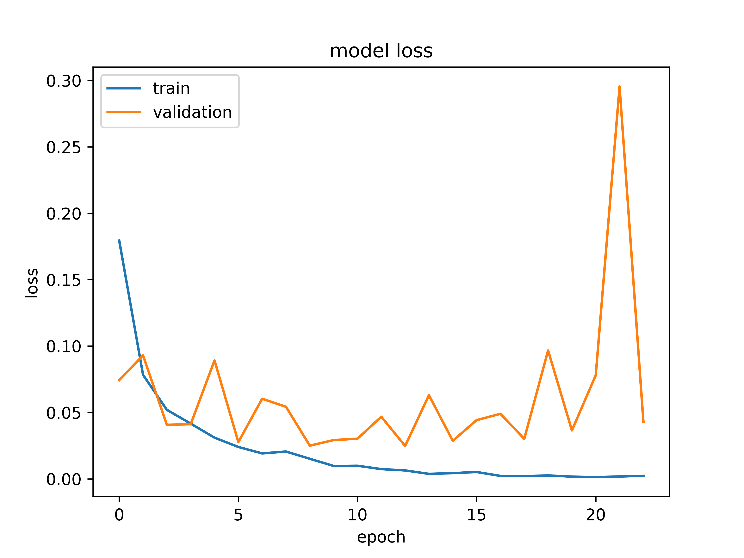}
\caption{Three-modality fusion training phase: train and validation accuracy (left) and loss (right).}
\label{fig:training}
\end{figure}

Early stopping halted training at epoch 12. Up to epoch~10 the validation accuracy rises together with the training accuracy, indicating a model that does not overfit; beyond that, accuracy plateaus while the validation loss becomes more volatile and trends upward (Fig.~\ref{fig:training}). This is precisely the condition early stopping is designed to detect --- the final model is taken before the loss deteriorates --- and the volatility itself largely reflects the limited validation-set size, where mini-batch composition perturbs the loss even as accuracy stays high and stable. Decisively, on the three held-out test recordings the three-modality model still achieves a 0.95 F1-score, demonstrating genuine generalization rather than memorization.

\subsection{Evaluation Metrics}
While sensor-fusion systems are often assessed on confidence, timeliness, throughput, and cost \cite{ref48}, this work addresses the offline binary classification problem and therefore adopts standard machine-learning metrics: Precision, Recall, Accuracy, the ROC curve, and F1-score. F1, the harmonic mean of Precision and Recall, is used as the primary metric for comparing the three models, as it is robust to the class imbalance caused by frequent false alarms (e.g., birds). Taking UAV as the positive class, Precision is the fraction of predicted UAVs that are correct and Recall (true-positive rate) is the fraction of actual UAVs recovered, while the ROC curve plots the true-positive against the false-positive rate across decision thresholds.

\section{Results}\label{sec:results}
The system was evaluated on three recordings that were preprocessed in the same way as the training set. The first recording was captured during the morning. The maximum range of the UAV from the sensors was 700 m while the UAV's maximum speed was 10 m/s. The second recording was captured during midday. The UAV's maximum range was 1 km and its maximum speed was 14 m/s. The third recording was captured in the morning and the UAV's maximum range from the sensor was 1km while its maximum speed was 11 m/s. In all three recordings the maximum altitude of the UAV was 80 m from the sensors level. Although the held-out test set comprises three recordings, these were deliberately selected to span diverse operating conditions --- different times of day (morning and midday), target ranges from 700 m to 1 km, target speeds between 10 and 14 m/s, and altitudes up to 80 m --- and each recording contributes many individual detection-level samples rather than a single test instance.

The recordings were mixed and shuffled (since the algorithm has no time dependencies) and produced a total number of 3209, 2662 and 1956 samples for the single modality (thermal), the two modalities (thermal-optronic) and the three modalities respectively. The reason for the different number of samples of each model, is the data registration procedure between the different modalities. The two modalities matched fewer data samples than the existing number of the thermal modality, while the three modalities matched even less. Because of the natural stochasticity of the deep neural networks (each execution may have a slightly different result), the models were trained five times and an average F1-score from each model was obtained. The results showed a 0.91 F1-score for the thermal modality alone, 0.93 F1-score for the two modalities while in the three modalities case, the model achieved a 0.95 average F1-score (Table~\ref{tab:f1}).

\begin{table}[htbp]
\caption{Classification results based on F1-score of the three models.}
\label{tab:f1}
\centering
\begin{tabular}{lccc}
\toprule
 & \textbf{Single} & \textbf{Two} & \textbf{Three} \\
 & \textbf{Modality} & \textbf{Modalities} & \textbf{Modalities} \\
\midrule
\#UAV samples & 1045 & 937 & 435 \\
\#FA samples  & 2164 & 1725 & 1521 \\
F1-score      & 0.91 & 0.93 & 0.95 \\
\bottomrule
\end{tabular}
\end{table}

\begin{figure}[t]
\centering
\includegraphics[width=\columnwidth]{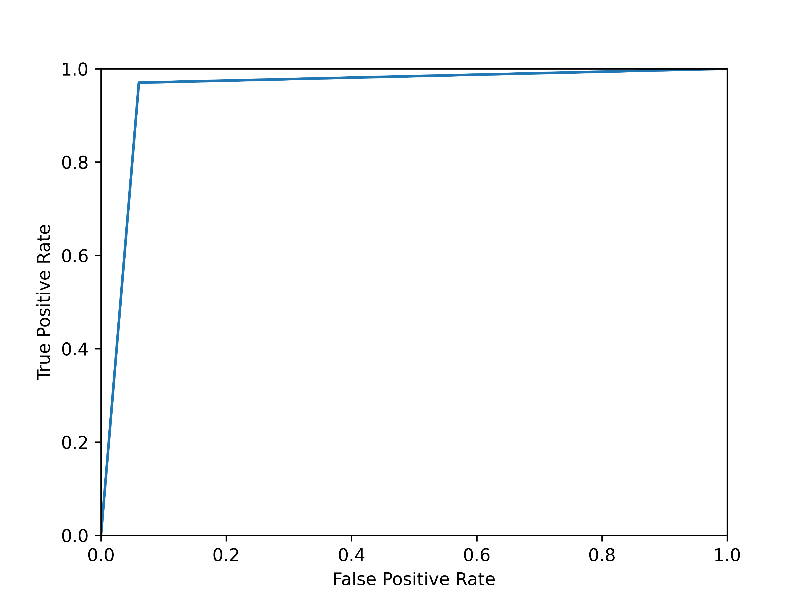}
\caption{Receiver Operating Characteristics (ROC) curve of the test set for the IR--Optronic--2D Radar data.}
\label{fig:roc}
\end{figure}

Both the two and three modality models achieve an increased performance compared to the single modality. Specifically, the three-sensor fusion model achieved the superior performance with a weighted average precision and recall of 0.94 and 0.94 respectively (Table~\ref{tab:prf}). Its confusion matrix is presented in Table~\ref{tab:cm} and its model's ROC curve is depicted in Fig.~\ref{fig:roc}. The residual errors are dominated by false positives: the UAV-class precision of 0.82 (Table~\ref{tab:prf}) shows that a minority of non-UAV airborne objects, such as birds and airplanes, are still misclassified as UAVs, which remains the principal error source for the fused model.

\begin{table}[htbp]
\caption{Three-modality fusion results for Precision, Recall and F1-score on the test set.}
\label{tab:prf}
\centering
\begin{tabular}{lccc}
\toprule
 & \textbf{Precision} & \textbf{Recall} & \textbf{F1-score} \\
\midrule
FA (0)        & 0.99 & 0.94 & 0.96 \\
UAV (1)       & 0.82 & 0.97 & 0.89 \\
Weighted avg  & 0.94 & 0.94 & 0.95 \\
\bottomrule
\end{tabular}
\end{table}

\begin{table}[htbp]
\caption{Confusion matrix of three-modality fusion.}
\label{tab:cm}
\centering
\begin{tabular}{lcc}
\toprule
 & \textbf{FA (0)} & \textbf{UAV (1)} \\
\midrule
FA (0)  & 1429 & 92 \\
UAV (1) & 13   & 422 \\
\bottomrule
\end{tabular}
\end{table}

The results show that fusing high-level feature maps from multiple sensors is an effective way to increase classification accuracy for the UAV-versus-false-alarm problem. A direct comparison with other multi-sensor methods is difficult, as no standardized full-system metrics exist and, to the authors' knowledge, no public dataset provides simultaneously recorded thermal, optronic, and 2D-radar UAV data. The controlled single- to multi-modality comparison reported here therefore serves as a fair internal benchmark under identical conditions, while releasing such a synchronized benchmark remains an important direction for future work.

The approach also has operational limitations. Feature-level fusion requires a registration step that matches data across modalities; detections that fail to match are discarded and, in a live system, could otherwise have supported single-sensor detection. A complete system should thus fall back to one- or two-sensor models when data are unavailable, and must account for the latency introduced by the one-second registration window, sensor communication, and inference --- factors that are critical given how quickly UAVs travel. Finally, the manual ground-truth labelling noted earlier also introduces annotator subjectivity and does not scale to large deployments, motivating semi-automatic or self-supervised labelling. A thorough field evaluation across attack scenarios --- measuring detection range and timeliness under $360^{\circ}$ coverage, analysing performance versus sensor-to-target distance \cite{ref28}, and extending the already-collected UAV-swarm recordings to dense-urban settings where sensor occlusion and interference may degrade performance --- is therefore left to future work.

\section{Conclusion}\label{sec:conclusion}
In this work a methodology to increase classification accuracy towards the UAV detection problem by fusing data from multiple ground sensors (thermal, optronic and 2D-radar) was presented. The employed algorithm utilized data spatial registration methods, feature stacking and CNNs, to fuse data from the sensors' individual DNN detection models into a new DNN fusion model to effectively increase the system's classification accuracy. The three-modality fusion model achieved an F1-score of 0.95, classifying more accurately UAVs from other objects in comparison to the two and the single modality models. The results demonstrate a consistent improvement as modalities are added, with the F1-score rising from 0.91 for the best single-sensor (thermal) baseline to 0.93 for the dual-modality (thermal--optronic) configuration and 0.95 for the full three-modality model, confirming that complementary sensor features carry information that the fusion network can exploit. The main contribution of this work lies in performing the fusion at the feature-map level across heterogeneous image and 2D-radar detectors, rather than at the input or decision level, while keeping the fusion head compact to suit the limited number of synchronized multi-sensor samples. The study also has a defined scope: the fusion is performed offline on pre-recorded data, and performance is reported on a held-out set from a single measurement campaign, since no public synchronized thermal--optronic--radar benchmark currently exists for direct comparison. These boundaries motivate the future directions discussed below --- releasing such a benchmark, evaluating the model across multiple independent campaigns, and validating the approach in live, on-edge operation. Future work will address the deployment of the fusion model on resource-constrained edge hardware, including the optimization of inference latency to meet the real-time requirements of operational counter-UAV systems.

\section*{Funding}
This research was funded by the European Union's Horizon 2020 Research and Innovation Program Advanced holistic Adverse Drone Detection, Identification and Neutralization (ALADDIN) [Grant Agreement No.~740859].

\end{document}